\documentclass[lettersize,journal]{IEEEtran}
\usepackage{amsmath,amsfonts}
\usepackage{algorithmic}
\usepackage{algorithm}
\usepackage{array}
\usepackage[caption=false,font=normalsize,labelfont=sf,textfont=sf]{subfig}
\usepackage{textcomp}
\usepackage{stfloats}
\usepackage{url}
\usepackage{verbatim}
\usepackage{graphicx}
\usepackage{cite}
\usepackage{color}
\usepackage{multirow}
\usepackage{booktabs}
\usepackage{float}
\usepackage{bm}
\usepackage[pagebackref,breaklinks,colorlinks,bookmarks=false]{hyperref}
\usepackage[T1]{fontenc}

\newcommand{\mc}[1]{\mathcal{#1}}

\def\eg{{\em e.g.}}
\def\ie{{\em i.e.}}
\def\etal{{\em et al. }}

\def\red#1{\textcolor{red}{#1}}
\def\blue#1{\textcolor{blue}{#1}}

\begin{document}
\title{Active Fake: DeepFake Camouflage}
\author{Pu Sun, Honggang Qi,~\IEEEmembership{Member,~IEEE},  Yuezun Li,~\IEEEmembership{Member,~IEEE}

\thanks{Honggang Qi and Yuezun Li are \textit{Corresponding authors.}}
\thanks{Pu Sun and Honggang Qi are with the University of Chinese Academy of Sciences, China. e-mail: (sunpu21@mails.ucas.ac.cn;hgqi@ucas.ac.cn).}
\thanks{Yuezun Li is with the School of Computer Science and
Technology, Ocean University of China, China. e-mail: (liyuezun@ouc.edu.cn). }
}

\markboth{Journal of \LaTeX\ Class Files,~Vol.~14, No.~8, August~2021}%
{Shell \MakeLowercase{\textit{et al.}}: A Sample Article Using IEEEtran.cls for IEEE Journals}


\maketitle

\begin{abstract}
DeepFake technology has gained significant attention due to its ability to manipulate facial attributes with high realism, raising serious societal concerns. Face-Swap DeepFake is the most harmful among these techniques, which fabricates behaviors by swapping original faces with synthesized ones. Existing forensic methods, primarily based on Deep Neural Networks (DNNs), effectively expose these manipulations and have become important authenticity indicators. However, these methods mainly concentrate on capturing the blending inconsistency in DeepFake faces, raising a new security issue, termed \textit{\blue{Active Fake}}, emerges when individuals intentionally create blending inconsistency in their authentic videos to evade responsibility. This tactic is called \textit{\blue{DeepFake Camouflage}}. To achieve this, we introduce a new framework for creating DeepFake camouflage that generates blending inconsistencies while ensuring \textit{imperceptibility}, \textit{effectiveness}, and \textit{transferability}. This framework, optimized via an adversarial learning strategy, crafts imperceptible yet effective inconsistencies to mislead forensic detectors. Extensive experiments demonstrate the effectiveness and robustness of our method, highlighting the need for further research in active fake detection.
\end{abstract}


\section{Introduction}
\begin{figure*}[t!]
    \centering
    \includegraphics[width=\linewidth]{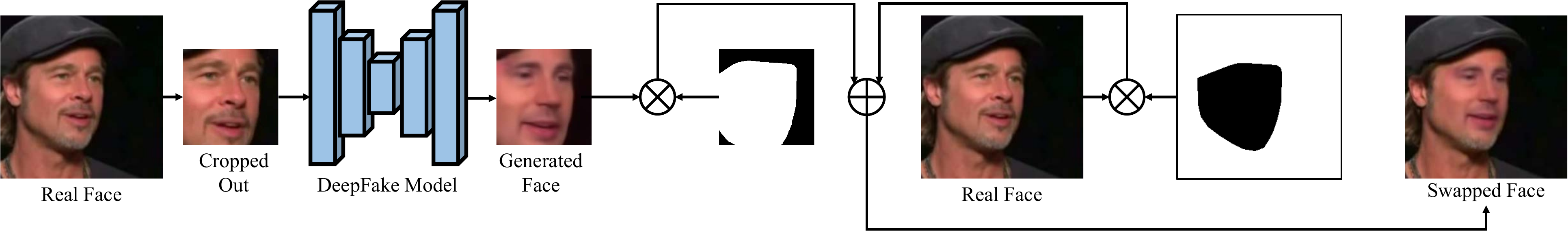}
    \caption{Pipeline of creating a Face-Swap DeepFake face. Firstly, the central face area is cropped out from the original face image, which is used to synthesize a target face. Secondly, this face is blended back to the original face image using a specific mask. }
    \label{fig:10_fsdf}
\end{figure*}

DeepFake is a recent AI generative technique that has drawn increasing attention. It manipulates facial attributes such as identity, expression, and movement with high realism, causing serious societal concerns, such as attacks on face recognition systems~\cite{akhtar2023deepfakes}, the spread of misinformations~\cite{akhtar2023deepfakes}, and threats to societal stability~\cite{pantserev2020malicious}. Among these techniques, Face-Swap DeepFake is particularly notable and harmful, as it can fabricate the behavior of target identities by swapping the original face with a synthesized target face~\cite{huang2023implicit,delfino2019pornographic,van2022deepfakes}. This technique has matured, and many Face-Swap tools have become prevalent and user-friendly, \eg, DeepFaceLab~\cite{liu2023deepfacelab}, Faceshifter~\cite{li2019faceshifter},  FaceSwap~\cite{faceswap}, Deepswap~\cite{deepswap}, and Faceswapper~\cite{faceswapper}. Two major steps are usually employed to create a Face-Swap DeepFake face: Firstly, the central face area is cropped out from the original face image, which is used to synthesize a target face. Secondly, this face is blended back to the original face image. Fig.~\ref{fig:10_fsdf} illustrates the pipeline of creating Face-Swap DeepFakes.
\textbf{It is important to note that the current DeepFake attacks belong to \textit{\red{Passive Fake}}, where attackers maliciously use photos of victims to create fake content without their consent.} 
These DeepFakes are visually indistinguishable from the naked eye, necessitating dedicated forensics methods to protect potential victims. 

The mainstream forensics methods are developed on Deep Neural Networks (DNNs), leveraging their powerful feature-capturing capacities to expose manipulation traces~\cite{rossler2019faceforensics++,afchar2018mesonet,  luo2021generalizing,zhao2021multi,wang2021representative,nataraj2019detecting,nguyen2019multi, li2020face, yin2024improving, guo2023exposing, chen2022self, nguyen2019capsule,yan2023ucf,liang2022exploring,liu2021spatial,qian2020thinking,  yu2023augmented, zhang2022unsupervised}. These methods have been demonstrated highly effective and are indispensable for verifying the authenticity of faces. 
Many of them are trained specifically on Face-Swap DeepFakes, allowing models to learn the specific manipulation patterns~\cite{rossler2019faceforensics++,afchar2018mesonet,luo2021generalizing,zhao2021multi,wang2021representative,nataraj2019detecting,nguyen2019multi,li2020face,  yin2024improving, guo2023exposing, chen2022self, nguyen2019capsule,yan2023ucf,liang2022exploring,liu2021spatial,qian2020thinking, yu2023augmented, zhang2022unsupervised}. Since blending operations are commonly used in creating Face-Swap DeepFakes, these manipulation patterns inherently contain blending inconsistency, \ie, the discrepancy between blended face area and authentic surroundings (validated in Sec.~\ref{sec:ins}).
Based on this, several advanced data augmentation strategies have been explored to improve the generalization of DeepFake detection. These methods typically create \textit{pseudo-fake faces} by blending various faces~\cite{li2020face, chen2022self}. This enables models to focus more on the inconsistency introduced by blending operations.

\begin{figure}[!t]
    \centering
    \includegraphics[width=0.85\linewidth]{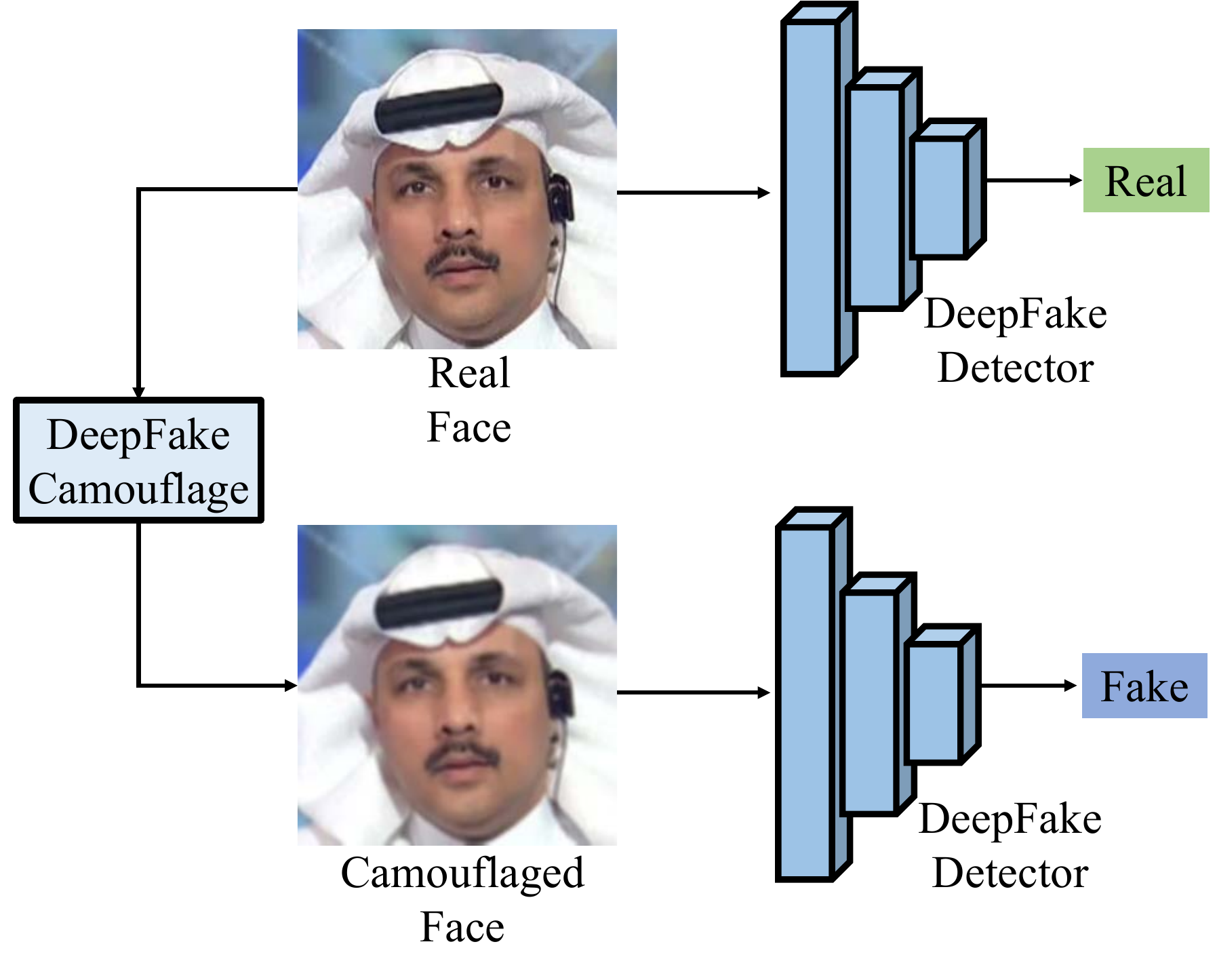}
    \caption{Overview of DeepFake camouflage.}
    \label{fig:00_intro}
\end{figure}

Despite their impressive performance, these methods face a new security problem that can be exploited maliciously, which we refer to as \textit{\textbf{\blue{Active Fake}}}. Since forensic methods rely on blending inconsistency as evidence of authenticity, individuals can intentionally create inconsistency in their authentic but inappropriate videos and release them publicly. If legislative institutions investigate and hold them accountable, they can falsely claim that these videos were manipulated by DeepFake. We refer to this tactic as 
\begin{center}
    \textit{\textbf{\blue{``DeepFake Camouflage''}}}  
\end{center}
Fig.~\ref{fig:00_intro} illustrates the idea of DeepFake camouflage.

Note that the adversarial attacks can also be used to mislead DNN models and have been explored in several anti-forensics methods for evading DeepFake detectors~\cite{gandhi2020adversarial,li2021exploring,hussain2021adversarial,jia2022exploring, hou2023evading}. However, these methods are limited in practical applications: 1) they are often difficult to interpret because they are typically generated by disrupting classification objectives without considering the context of Face-Swap DeepFakes. 2) they are visible since they are not related to the face content. 3) they concentrate on specific models rather than the essence of DeepFake detection, making them more overfitted to specific models (See Sec.~\ref{sec:results}).

In this paper, we break away from previous methods and introduce a new framework for DeepFake camouflage. Our idea is to apply simple image operations with learned parameters on real faces to introduce blending inconsistency while satisfying three key criteria: 1) \textit{imperceptibility} to human observers; 2) \textit{effectiveness} in deceiving the DeepFake detectors; 3) \textit{transferability} across various mainstream DeepFake detectors.  

To achieve this, we propose \textit{Camouflage GAN (CamGAN)}, a framework designed to generate blending inconsistency that evades DeepFake detectors. Our framework comprises four key components: a configuration generator, a camouflage module, a visual discriminator, and a DeepFake detector (See Fig. \ref{fig:03_workflow}). Specifically, we employ two operations to create inconsistency, Gaussian noising and Gaussian filtering. The configuration generator determines the intensity of these operations by learning to craft the appropriate parameters based on the input image. 
The camouflage module preprocesses the face area with these learned parameters and blends it into the original face image using a blending mask derived from facial landmarks. It enhances visual quality by minimizing artifacts around the blending boundary through Gaussian filtering, with parameters generated by the configuration generator. 
Training the configuration generator involves adversarial learning~\cite{goodfellow2014generative} with two discriminators: a visual discriminator, which ensures the camouflaged face appears visually real, and a DeepFake detector, which validates if the camouflaged face can mislead the detector. This training process is challenging as the operations within the camouflage module are not differentiable, making gradient back-propagation inapplicable. Thus, we describe a reinforcement learning-based scheme to optimize the framework. During inference, only the configuration generator and the camouflage module are needed to create camouflaged faces. 


Our contributions are summarized as follows:
\begin{itemize}
    \item We introduce a new approach, \textit{DeepFake Camouflage}, to evade DeepFake detectors. Unlike conventional passive fake methods, this approach allows attackers to release authentic but inappropriate videos publicly while avoiding accountability. 
    
    \item  We propose a new generative framework (CamGAN) to achieve DeepFake Camouflage. CamGAN learns to generate appropriate preprocessing parameters to create blending inconsistency on authentic face images. This framework is adversarially trained and optimized using a reinforcement learning mechanism.
    
    \item  Extensive experiments on standard datasets, in comparison to various adversarial attacks, demonstrate the effectiveness of our method in multiple scenarios. We also thoroughly study the effect of each component, offering insights for future research in active fake.
\end{itemize}

\section{Background and Related Work}
\subsection{DeepFake}
DeepFake, a combination of Deep Learning and Fake Face, first appeared on Reddit in 2017\cite{nguyen2019deep}. Originally, DeepFake referred to a face-swap technique capable of generating highly realistic target identity faces and replacing the source identity faces in videos while maintaining consistent facial attributes such as expressions and orientation, as illustrated in Fig. \ref{fig:10_fsdf}. Nowadays, the term DeepFake has expanded to encompass all AI-generated faces, including whole face synthesis (created by GANs~\cite{heusel2017gans,karras2017progressive,karras2019style} and diffusion models~\cite{kim2023dcface,stypulkowski2024diffused}), face editing~\cite{huang2023collaborative,xu2022transeditor,jo2019sc}, and face reenactment~\cite{hsu2024pose,yang2022face2face,bounareli2023hyperreenact}. Nevertheless, among these forms, Face-Swap DeepFake has gained the most attention due to its significant negative social impacts, such as the creation of revenge porn videos~\cite{delfino2019pornographic}, the fabrication of inappropriate behavior by public figures~\cite{van2022deepfakes}, and economic fraud~\cite{van2022deepfakes}. The availability of user-friendly tools for Face-Swap DeepFakes has further lowered the barrier to making fake videos, thereby exacerbating the security risks. Therefore, this paper focuses specifically on the Face-Swap DeepFakes. 

\subsection{DeepFake Detection}
To curb the misuse of DeepFake algorithms, DeepFake Detection algorithms have flourished in recent years. DeepFake Detection aims to perform binary classification on input images or videos to determine their authenticity. They could be roughly categorized as naive, spatial and frquency detectors~\cite{yan2024deepfakebench}. Many CNN-based models are utilized in DeepFake Detection,  with data-driven training~\cite{rossler2019faceforensics++} and various strategies, such as new designed architectures~\cite{afchar2018mesonet,luo2021generalizing,zhao2021multi, zhang2022unsupervised}, augmentations~\cite{wang2021representative, yu2023augmented, yin2024improving} and preprocessing~\cite{nataraj2019detecting}.  ~\cite{li2020face, chen2022self} create pseudo-fake faces by blending different faces as s special data augmentation. Some DeepFake detectors specifically utilize the spatial information of images~\cite{nguyen2019multi, nguyen2019capsule, yan2023ucf,liang2022exploring}.~\cite{yan2023ucf,liang2022exploring} introduce disentanglement learning into DeepFake detection. Nguyen \etal~\cite{nguyen2019multi} locates the forgery region besides classify the image.~\cite{nguyen2019capsule} utilize capsule network to detect the images. Some other detectors fully utilize the information in  frequency domain for detection~\cite{qian2020thinking,liu2021spatial,luo2021generalizing}. Qian  \etal~\cite{qian2020thinking} propose to learn the subtle forgery patterns through frequency components partition. SPSL~\cite{liu2021spatial} utilize phase spectrum analysis to improve the classification. SRM~\cite{luo2021generalizing} notice the high-frequency noise could boost the performance. To fully prove the effectiveness of our method, we perform experiments on naive, spatial, and frequency detectors.

\subsection{Evading DeepFake Detection}
Existing methods for evading DeepFake detection~\cite{gandhi2020adversarial,li2021exploring,hussain2021adversarial,jia2022exploring, hou2023evading} commonly use adversarial attacks~\cite{goodfellow2014explaining} to add noise to the images, misleading DeepFake detectors to make incorrect predictions. While these approaches have shown promise, they suffer from several significant limitations that hinder their practical application: 1) Poor interpretability: Adversarial perturbations are typically generated by disrupting classification objectives and back-propagating gradients to the input face image. These perturbations often have little connection to the context of Face-Swap DeepFakes, making them difficult to interpret. 2) High visibility: The visibility of adversarial perturbations is closely tied to the content of the face image. However, because these attacks are designed without considering the specific facial content, they often result in highly visible artifacts. 3) Limited transferability: Adversarial attacks generally focus on targeting specific models, which can lead to overfitting and poor transferability to other models. Despite attempts to address this issue, this limitation is inherent because the design of these attacks does not align with the fundamental nature of DeepFake detection. Therefore, we depart from these existing methods and introduce a new framework to achieve DeepFake Camouflage.

\section{Active Fake: DeepFake Camouflage}
\label{sec:camops}

\label{sec:preval}
\begin{figure}[!t]
    \centering
    \includegraphics[width=\linewidth]{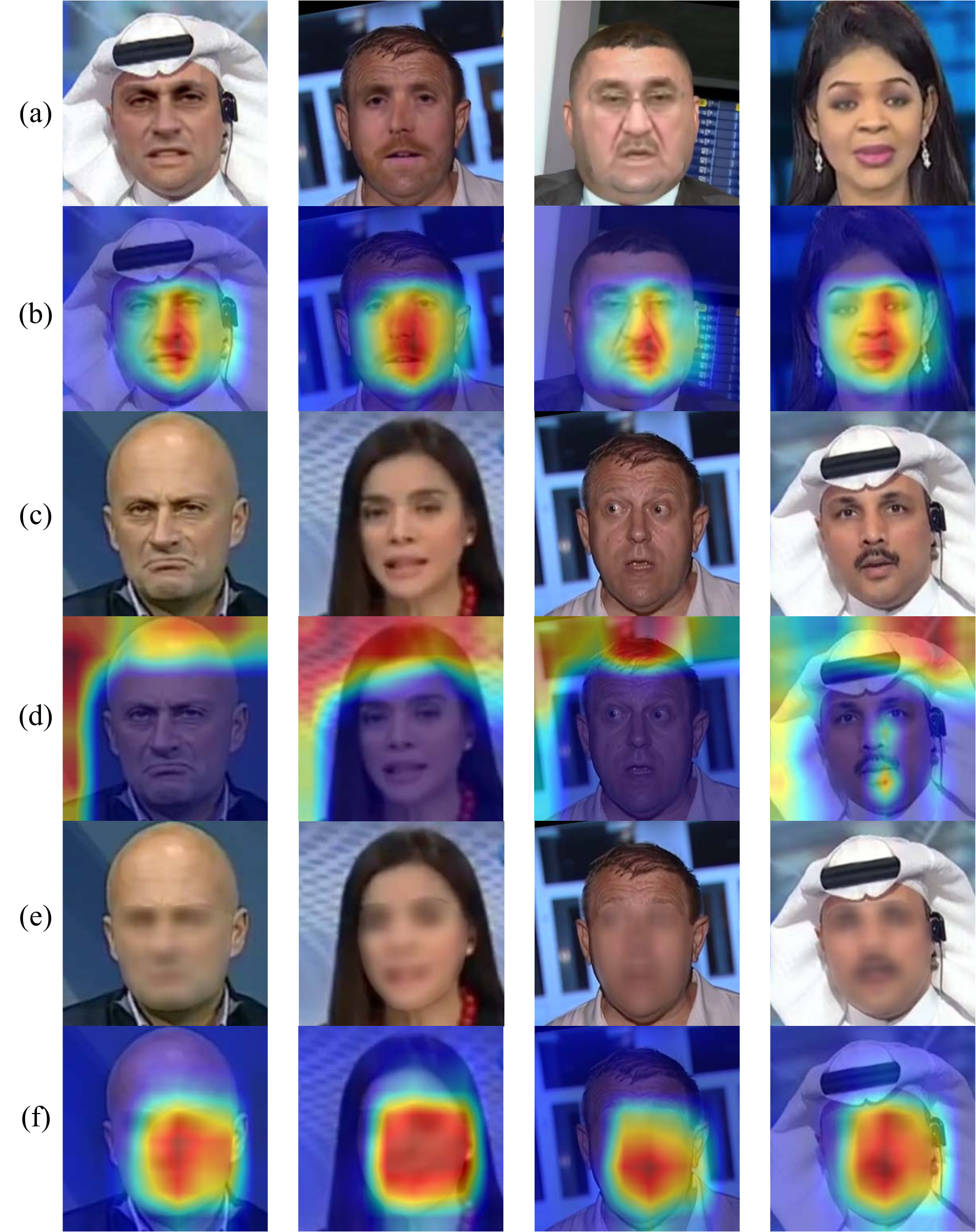}
    \caption{Grad-CAM maps on different  images. Row (a) \& (b): DeepFake images and their Grad-CAM maps. Row(c) \& (d):Real clean images and their Grad-CAM maps. Row(e) \& (f): Real images with handcrafted artifacts and their Grad-CAM maps.}
    \label{fig:06_preval}
\end{figure}

\subsection{Inspiration and Preliminary Analysis}
\label{sec:ins}
To verify the feasibility of our idea, we employ Xception~\cite{chollet2017xception} as the DeepFake detector and train it using Face-Swap DeepFake faces. As shown in Fig.~\ref{fig:06_preval}, we visualize the attention of Xception on DeepFake faces using Grad-CAM~\cite{selvaraju2017grad}. Row (a) indicates the DeepFake faces and Row (b) exhibits corresponding Grad-CAM maps. These visualizations demonstrate that the detector mainly concentrates on the manipulated face area. In contrast, the real face images, as shown in Row (c, d), have scattered attention over the backgrounds. 

To demonstrate whether the detectors treat the blending inconsistency as important evidence of authenticity, we manually create an inconsistency by manually applying intense Gaussian filtering to the central face area of real images. The visual examples are shown in Row (e). We send these images into the detector and visualize the Grad-CAM maps. As shown in Row (f), these images are successfully identified as fake and are highlighted on the processed face area as in Row (b), demonstrating the feasibility of disrupting the detectors by introducing inconsistency. 

\textbf{Nevertheless, manually designing the inconsistency is infeasible, as it can hardly maintain imperceptibility, effectiveness, and transferability. Thus we describe a learnable framework to create blending inconsistency.}




\subsection{Problem Setup \& Overview}
\label{sec: setup}
Denote $\bm{x}_r \in \mathbb{R}^{H \times W \times 3}$ as a real clean face image and $\bm{x}^*_r$ as the camouflaged face image. Let a DeepFake detector as $\bm{D}$. Our goal is to create imperceptible blending inconsistency on real faces, causing them to be classified as fake. This goal can be written as
\begin{equation}
        \min_{\bm{w}} \; \lVert \bm{x}^*_r -  \bm{x}_r \rVert_{p}, \; 
        \textrm{s.t.} \; \bm{D} \left (\bm{x}^*_r \right ) = 0,
\end{equation}
where $\lVert \cdot \rVert$ indicates the magnitude of the inconsistency, $\{0, 1\}$ represents fake and real, respectively.

Denote the camouflage module as $\bm{C}$. The camouflaged face can be denoted as $\bm{x}^*_r = \bm{C}(\bm{x}_r; \bm{w})$, where $\bm{w}$ represents the parameters learned in process. The camouflage module $\bm{C}$
involves two steps: creating inconsistency and blending inconsistency. 

\smallskip
\noindent\textbf{Creating Inconsistency.}
To create inconsistency, we adopt two image operations: Gaussian noising and Gaussian filtering. Denote the parameters for Gaussian noising as $w_{\textrm{gn}} = (\mu_{\textrm{gn}}, \sigma_{\textrm{gn}})$, where $\mu_{\textrm{gn}}, \sigma_{\textrm{gn}}$ correspond to the mean and standard variance. 
After adding Gaussian noise, we then apply Gaussian filtering. This operation blurs the images using a Gaussian kernel. The parameters for Gaussian filtering is denoted as $w_{\textrm{gf}} = (k_{\textrm{gf}}, \sigma_{\textrm{gf}})$, where $k_{\textrm{gf}}, \sigma_{\textrm{gf}}$ correspond to the kernel size and standard variance. Denote $\bm{x}'_{r}$ as the face image after adding inconsistency.

\smallskip
\noindent\textbf{Blending Inconsistency.}
\label{sec:adb}
We blend the facial region of $\bm{x}'_{r}$ into the original face image $\bm{x}_{r}$ using a mask $\mathcal{M}$. This process can be described as
\begin{equation}
    \bm{x}^*_{r} = \bm{x}'_{r} \cdot \mathcal{M} + \bm{x}_{r} \cdot (1 - \mathcal{M}).
    \label{eq:blend}
\end{equation}
Straightforwardly, the mask $\mathcal{M}$ is the convex hull that includes all the facial contour landmarks, where the pixels inside the mask are set to $1$ and those outside the mask are set to $0$ (see Fig.\ref{fig:10_fsdf}). 
However, simply using this mask can introduce visible artifacts around the blending boundary, due to the color or texture discrepancy between $\bm{x}'_{r}$ and $\bm{x}_{r}$. Inspired by~\cite{li2020face}, we convert the binary mask into a soft mask by applying Gaussian filtering on mask boundary and use it for face blending. Denote the parameters for Gaussian filtering on mask as $w_{\textrm{bl}} = (k_{\textrm{bl}}, \sigma_{\textrm{bl}})$. Note that these parameters are also learned.






\begin{figure*}[!t]
    \centering
    \includegraphics[width=\linewidth]{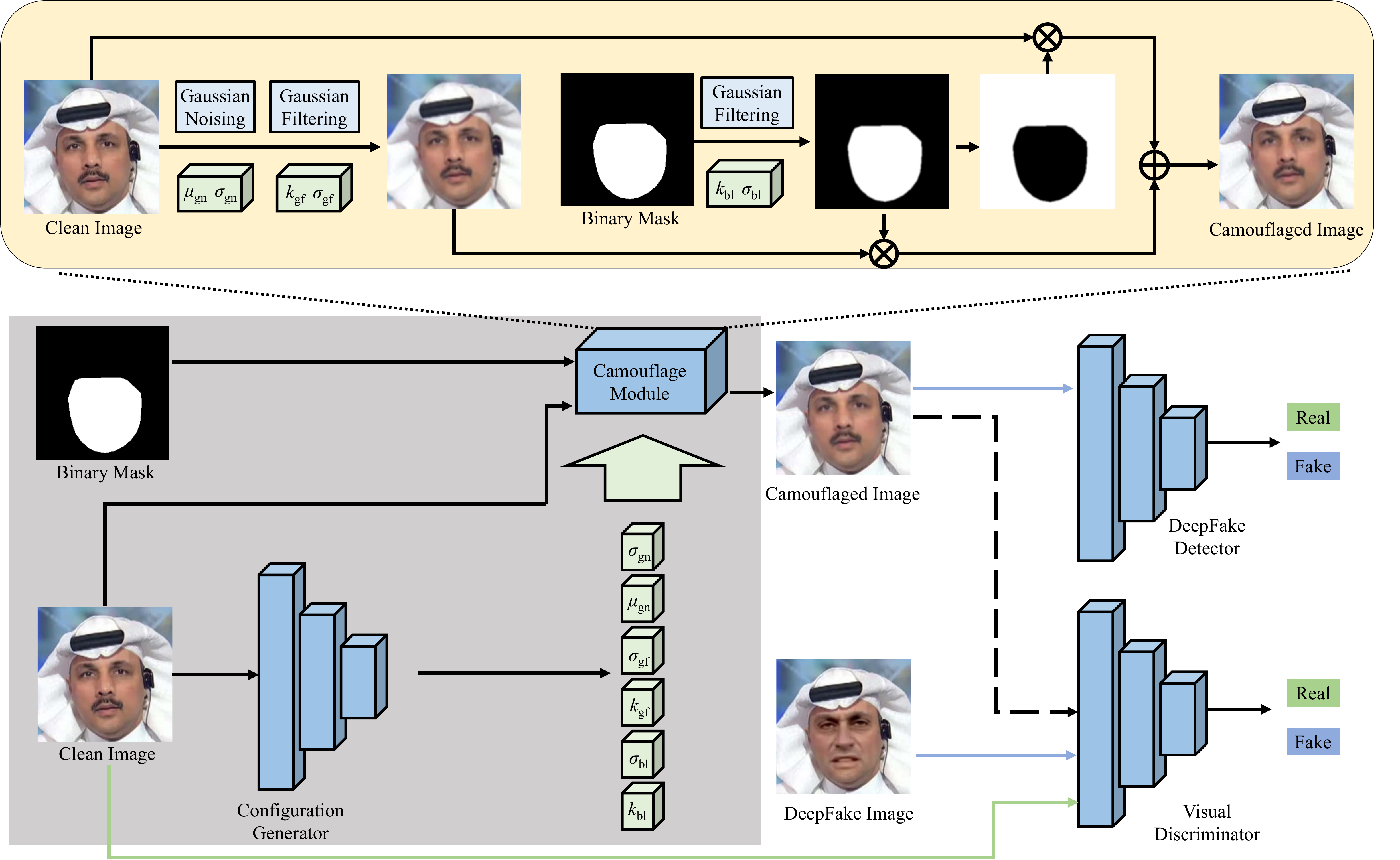}
    \caption{Training pipeline of Camouflage GAN. The inference pipeline is in the gray background area. The green (real) and blue (fake) arrows  represent the expected network output when the corresponding images are input in the training phase. The dashed arrows indicate that, when optimizing the configuration generator, the goal output of the  visual discriminator taking camouflaged images as input should be Real; When optimizing the visual discriminator, the goal output of the visual discriminator taking camouflaged images as input should be Fake;  See Sec.~\ref{sec:CamGAN} for details.}
    \label{fig:03_workflow}
\end{figure*}

\subsection{Camouflage GAN}
\label{sec:CamGAN}

To determine the values of parameters $\bm{w} = (w_{\textrm{gn}},w_{\textrm{gf}},w_{\textrm{bl}})$, we propose a Camouflage GAN (\textit{CamGAN}) that learns to generate parameters adaptive to the different face images.

\smallskip
\noindent\textbf{Overview and Architectures.}
As shown in Fig.~\ref{fig:03_workflow}, this framework is composed of four key components: a configuration generator $\bm{G}$, a camouflage module $\bm{C}$, a visual discriminator $\bm{V}$, and a DeepFake detector $\bm{D}$. 
\begin{itemize}
    \item[-] \textit{Configuration Generator.} This generator is designed to create all learnable parameters $\bm{w}= (\sigma_{\textrm{gn}}$, $\mu_{\textrm{gn}}$, $\sigma_{\textrm{gf}}$, $k_{\textrm{gf}}$, $\sigma_{\textrm{bl}}$, $k_{\textrm{bl}})$. This generator is developed on Xception~\cite{rossler2019faceforensics++} with six additional parallel fully connected layers for predicting corresponding parameters. 

    \item[-] \textit{Camouflage Module.} Given a clean real face $\bm{x}_r$, we first apply Gaussian noising and Gaussian filtering with the learned parameters $(\sigma_{\textrm{gn}}$, $\mu_{\textrm{gn}}$, $\sigma_{\textrm{gf}}$, $k_{\textrm{gf}})$. Then we create a blending mask $\mathcal{M}$ using the following steps: {We first obtain a binary mask by drawing a convex hull including all facial boundary landmarks. We then apply Gaussian filtering with the learned parameters $\sigma_{\textrm{bl}}$, $k_{\textrm{bl}}$  to this binary mask to obtain a soft mask as blending mask $\mathcal{M}$. The whole process is shown in Fig.~\ref{fig:03_workflow} (yellow box).}
    
    \item[-]\textit{Visual Discriminator.} This discriminator is designed to simulate the human eyes, distinguishing between images with inconsistency and not. We also employ the Xception network and output a binary classification, \ie, whether the input face having inconsistency. The camouflaged face $\bm{x}^{*}_{r}$ and DeepFake face $\bm{x}_{f}$ are expected to have inconsistency, while real images $\bm{x}_{r}$ are not. 

    \item[-] \textit{DeepFake Detector.} This detector serves as a discriminator for distinguishing whether a face is real or fake. Note that the camouflaged faces should be detected as fake. In our method, we directly employ the well-trained DeepFake detectors.

\end{itemize}

\subsection{Loss functions and Training}

Denote the configuration generator, the visual discriminator, and the DeepFake detector as $\bm{G}$, $\bm{V}$, and $\bm{D}$, respectively. To effectively instruct the learning of CamGAN, we introduce three simple loss terms: detector spoofing loss $\mathcal{L}_{\rm ds}$, visual inspection loss $\mathcal{L}_{\rm vi}$, and visual constraint loss $\mathcal{L}_{\rm vc}$.

\smallskip
\noindent\textbf{Detector Spoofing Loss.}
We expect that the camouflaged face $\bm{x}_{r}^{*}$ should be able to spoof the DeepFake detector $\bm{D}$, \ie, misleading the prediction of $\bm{x}_{r}^{*}$ as fake. Denote $0$ and $1$ correspond to fake and real respectively. Therefore, the detector spoofing loss can be defined as
\begin{equation}
\mc{L}_{\rm ds} = \log \bm{D}(\bm{x}_{r}^{*}),
\label{eq:spoof}
\end{equation}
where $\bm{D}(\bm{x}_{r}^{*})$ represents the probability of $\bm{x}_{r}^{*}$ being real. Minimizing this loss term can decrease the real probability of $\bm{x}_{r}^{*}$, \ie, being closer to label $0$. Note that we directly employ well-trained DeepFake detectors and fix their parameters during training.

\smallskip
\noindent\textbf{Visual Inspection Loss.}
The visual discriminator $\bm{V}$ is designed to determine whether the given face is visually manipulated. We employ this discriminator to improve the synthesized quality in a way of adversarial learning. Specifically, given the camouflaged face $\bm{x}_{r}^{*}$, we anticipate it can mislead this discriminator $\bm{V}$, \ie, being classified as real. Denote $y \in \{0, 1\}$ correspond to the label of visually fake and real respectively. This loss term can be defined as
\begin{equation}
\mc{L}_{\rm vi} =  -y \log \bm{V}(\bm{x}_{r}) + (1 - y) \log \bm{V}(\bm{x}_{f}) + \log \bm{V}(\bm{x}_{r}^{*}),
\label{eq:dis}    
\end{equation}
where $\bm{x}_{r}$ and $\bm{x}_{f}$ denote the wild real and fake faces, and $\bm{V}(\cdot)$ represents the probability of input face being real.

\smallskip
\noindent\textbf{Visual Constraint Loss.}
To ensure the camouflaged faces are visually similar to real faces, we design a visual constraint loss to restrict the strength of distortions. This loss term can be formulated as the $\ell_p$ norm distance between $\bm{x}_{r}^{*}$ and $\bm{x}_{r}$, as
\begin{equation}
      \mathcal{L}_{\rm vc} = \lVert \bm{x}_{r}^{*} - \bm{x}_{r} \rVert_{p},
\end{equation}

\smallskip
\noindent\textbf{Overall Loss and Optimization.} With these loss terms, we train CamGAN in the way of adversarial learning, which is expressed as 
\begin{equation}
        \min_{G} \max_{V} \; 
        \mc{L}_{\rm ds} + \mc{L}_{\rm vc} - \mc{L}_{\rm vi}.
    \label{eq:overall}
\end{equation}
Note that $\mc{L}_{\rm ds}, \mc{L}_{\rm vc}$ only involve optimizing the configuration generator $\bm{G}$, while $\mc{L}_{\rm vi}$ involves optimizing the configuration generator $\bm{G}$ and the visual discriminator $\bm{V}$. We employ the scheme of adversarial training, which alternately optimizes $\bm{G},\bm{V}$.
\begin{itemize}
    \item When fixing the configuration generator $\bm{G}$, both $\mc{L}_{\rm ds}$ and $\mc{L}_{\rm vc}$ remain unchanged. During this process, we maximize $-\mc{L}_{\rm vi}$, leading to the reduction of $\bm{V}(\bm{x}_{r}^{*})$, corresponding to classify $\bm{x}_{r}^{*}$ as fake. 

    \item When fixing discriminator $\bm{V}$, we minimize $\mc{L}_{\rm ds} + \mc{L}_{\rm vc} - \mc{L}_{\rm vi}$ by optimizing the generator $\bm{G}$. This means that the camouflaged faces aim to 1) spoof the DeepFake detector $\bm{D}$, \ie, being classified as fake, 2) have minimal distortions, and 3) deceive the visual discriminator $\bm{V}$, \ie, being classified as real.
\end{itemize}

\smallskip
\noindent\textbf{Reinforcement Learning Based Optimization.}
It is important to note that the process in the face synthesizer is typically not differentiable, leading to \textit{gradient interruption}. This challenge affects the optimization of the configuration generator $\bm{G}$, preventing it from being optimized by standard gradient back-propagation. To resolve this, we adopt the strategy in reinforcement learning~\cite{williams1992simple} to optimize the configuration generator $\bm{G}$. 

Specifically, we reformulate the visual constraint loss $\mc{L}_{\rm vc}$ by disconnecting the configuration generator $\bm{G}$ with the face synthesizer and directly restricting the output of the configuration generator $\bm{G}$. Since the generated operation parameters control the magnitude of face distortion, restricting them helps ensure the visual similarity between the camouflaged faces and real faces. 

The configuration generator $\bm{G}$ is crafted to output three sets of operation parameters as $\bm{w} = (w_{\rm gn}, w_{\rm gf}, w_{\rm bl})$, corresponding to the Gaussian noising and Gaussian
filtering in \textit{creating inconsistency} step and the Gaussian masking in \textit{blending inconsistency} step. The larger magnitude of $w_{\rm gn}, w_{\rm gf}$ introduces more intense distortions. On the contrary, the smaller magnitude of $w_{\rm bl}$ denotes the boundary of the blending mask is sharper, introducing more blending artifacts. Therefore, the visual constraint loss is reformulated to enlarge $w_{\rm gn}, w_{\rm gf}$ while restricting $w_{\rm bl}$, as expressed by
\begin{equation}
      \mathcal{L}_{\rm vc} = \log (w_{\rm gf}) + \log (w_{\rm gn}) - \log (w_{\rm bl}).
\end{equation}
Minimizing this equation corresponds to the better visual quality of camouflaged faces. 
We then optimize $\bm{G}$ using the following equation as
\begin{equation}
\theta_{t+1} = \theta_t - \eta \cdot (e^{\phi(\mc{L}_{\rm ds})} - \lambda \cdot  e^{\phi(\mc{L}_{\rm vi})}) \cdot  \nabla_{\theta_t} \mathcal{L}_{\rm vc},
\label{eq:rl_cg}
\end{equation}
where $\eta$ is an optimization step, $e^{\phi(\mc{L}_{\rm ds})} - \lambda \cdot  e^{\phi(\mc{L}_{\rm vi})}$ is the penalty term,  $\lambda$ is a weighting coefficient used to dynamically adjust the weights of the two losses during training, and $\phi$ is the $\texttt{sigmoid}$ function defined as
\begin{equation}
\phi(\mc{L}) = \frac{1}{1 + e^{-\mc{L}}}
\label{eq:rl_cg1}
\end{equation}
This approach allows the penalty term $e^{\phi(\mc{L}_{\rm ds})} - \lambda \cdot  e^{\phi(\mc{L}_{\rm vi})}$  to influence the parameters of $\bm{G}$ by optimizing Eq.\eqref{eq:rl_cg}, thereby approximating the process described in Eq.\eqref{eq:overall}.




 

\section{Experiments}

\subsection{Experimental Setups}

\smallskip
\noindent\textbf{Datasets.}
We use the training set of FaceForensics++~\cite{rossler2019faceforensics++} to train our model. In the testing phase, for FaceForensics++~\cite{rossler2019faceforensics++} dataset, we directly use the real faces in its testing set. Since there is no division into training and testing sets for the Celeb-DF~\cite{li2020celeb} dataset, we choose the real faces of 10 identities as the testing set. All the faces have the size of $256 \times 256$ as DeepFakeBench~\cite{yan2024deepfakebench}. 

\smallskip
\noindent\textbf{Metrics.}
We use four metrics for evaluation: ACC, SSIM, PSNR, and FID.
(1)ACC is the accuracy of the DeepFake detector in predicting whether an image is real or fake. We calculate the accuracy of various DeepFake detectors on real clean images and their corresponding camouflaged images. The greater the decrease in accuracy, the stronger the misleading effect our method has on the DeepFake detector. (2) SSIM, PSNR, and FID~\cite{heusel2017gans} are calculated between real clean images and their corresponding camouflaged images. They are used to measure the quality of camouflaged face images and how much our method has impacted the quality of the images. Higher SSIM and PSNR values, along with lower FID values, indicate smaller noise added to the camouflaged images.

\smallskip
\noindent\textbf{Implementation Details.}
We train the DeepFake detectors, Xception \cite{rossler2019faceforensics++}, FFD \cite{dang2020detection}, SPSL \cite{liu2021spatial}, and SRM \cite{luo2021generalizing}, using DeepfakeBench~\cite{yan2024deepfakebench} with FaceForensics++~\cite{rossler2019faceforensics++} dataset and fix their parameters as well-trained DeepFake detectors for all the subsequent experiments. These DeepFake detectors encompass naive, spatial, and frequency detectors.  Our method for Camouflage GAN is implemented using PyTorch 1.9.0 on Ubuntu 20.04 with an Nvidia 3090 GPU. In the experiments, the batch size is set as $1$, for optimizer we utilize RMSProp optimizer~\cite{hinton2012neural}, the initial learning rate is set as $1.0\times10^{-5}$. When adding Gaussian noise to an image, we first normalize the image to the range of $0.0$ to $1.0$ by dividing $255.0$. After adding Gaussian noise, we then multiply the image by $255.0$.

\subsection{Results}
\label{sec:results}

\begin{table}[!t]
\centering
\caption{ACC on FaceForensics++ dataset. No Attack represents real clean images.}\label{tab:ff-acc}
\begin{tabular}{ l | c | c | c | c }
\toprule

Attacks & Xception  & FFD  & SPSL & SRM\\
\midrule
\midrule

No Attack & 0.87 & 0.94 & 0.77 & 0.87 \\
\midrule
\midrule
Ours-Xception & 0.01 & 0.00 & 0.07 & 0.00 \\
Ours-FFD & 0.10 & 0.03 & 0.19 & 0.04 \\
Ours-SPSL & 0.18 & 0.03 & 0.24 & 0.05 \\
Ours-SRM & 0.14 & 0.02 & 0.23 & 0.03 \\
\midrule
CW-Xception   & 0.01 & 0.36 & 0.23 & 0.32 \\
CW-FFD        & 0.23 & 0.01 & 0.46 & 0.12 \\
CW-SPSL       & 0.11 & 0.41 & 0.14 & 0.36 \\
CW-SRM        & 0.51 & 0.33 & 0.63 & 0.03 \\
\midrule
Jitter-Xception & 0.34 & 0.85 & 0.66 & 0.97 \\
Jitter-FFD      & 0.69 & 0.31 & 0.76 & 0.91 \\
Jitter-SPSL     & 0.37 & 0.93 & 0.46 & 1.00 \\
Jitter-SRM      & 0.65 & 0.78 & 0.77 & 0.57 \\
\midrule
PGD-Xception    & 0.00 & 0.18 & 0.05 & 0.79 \\
PGD-FFD         & 0.01 & 0.00 & 0.34 & 0.01 \\
PGD-SPSL        & 0.00 & 0.69 & 0.01 & 0.98 \\
PGD-SRM         & 0.19 & 0.01 & 0.59 & 0.00 \\
\midrule
Pixle-Xception  & 0.13 & 0.48 & 0.67 & 0.51 \\
Pixle-FFD       & 0.83 & 0.03 & 0.75 & 0.27 \\
Pixle-SPSL      & 0.72 & 0.48 & 0.59 & 0.48 \\
Pixle-SRM       & 0.83 & 0.42 & 0.75 & 0.05 \\
\bottomrule
\end{tabular}
\end{table}

\begin{table}[htbp]
\centering
\caption{ACC on Celeb-DF dataset. No Attack represents real clean images.}\label{tab:celeb-acc}
\begin{tabular}{ l | c | c | c | c }
\toprule

Attacks & Xception  & FFD  & SPSL & SRM\\

\midrule
\midrule

No Attack & 0.78 & 0.69 & 0.58 & 0.52\\
\midrule
\midrule
Ours-Xception & 0.01 & 0.00 & 0.04 & 0.00 \\
Ours-FFD & 0.14 & 0.00 & 0.13 & 0.01 \\
Ours-SPSL & 0.35 & 0.03 & 0.20 & 0.10 \\
Ours-SRM & 0.38 & 0.06 & 0.24 & 0.16 \\
\midrule
CW-Xception   & 0.01 & 0.16 & 0.13 & 0.16 \\
CW-FFD        & 0.25 & 0.01 & 0.34 & 0.07 \\
CW-SPSL       & 0.19 & 0.25 & 0.10 & 0.23 \\
CW-SRM        & 0.45 & 0.24 & 0.45 & 0.01 \\
\midrule
Jitter-Xception & 0.30 & 0.78 & 0.51 & 0.97 \\
Jitter-FFD    & 0.61 & 0.26 & 0.55 & 0.93 \\
Jitter-SPSL   & 0.27 & 0.87 & 0.33 & 1.00 \\
Jitter-SRM    & 0.61 & 0.65 & 0.59 & 0.57 \\
\midrule
PGD-Xception  & 0.00 & 0.06 & 0.02 & 0.80 \\
PGD-FFD       & 0.01 & 0.00 & 0.25 & 0.11 \\
PGD-SPSL      & 0.00 & 0.51 & 0.01 & 0.98 \\
PGD-SRM       & 0.19 & 0.00 & 0.41 & 0.00 \\
\midrule
Pixle-Xception & 0.08 & 0.22 & 0.41 & 0.18 \\
Pixle-FFD     & 0.63 & 0.02 & 0.50 & 0.07 \\
Pixle-SPSL    & 0.54 & 0.20 & 0.39 & 0.14 \\
Pixle-SRM     & 0.65 & 0.22 & 0.51 & 0.01 \\
\bottomrule
\end{tabular}
\end{table}

\smallskip
\noindent\textbf{Quantitative Results.}
We conduct quantitative evaluations on our method as well as four adversarial attacks, \ie, CW~\cite{carlini2017towards}, Jitter~\cite{schwinn2023exploring}, PGD~\cite{madry2017towards}, and Pixle~\cite{pomponi2022pixle}, and the results are presented in Table \ref{tab:ff-acc}, Table \ref{tab:ff-ssim}, Table \ref{tab:celeb-acc}, and Table \ref{tab:celeb-ssim}. The left column in each table denotes the attack methods, \eg, Ours-FFD represents our CamGAN trained with FFD as the DeepFake detector, and CW-FFD represents using CW to attack FFD. The top row of  Table \ref{tab:ff-acc} and Table \ref{tab:celeb-acc} denotes which DeepFake detector is used in testing. It can be observed that our method outperforms those four adversarial attacks in terms of both white-box\footnote{The DeepFake detector being attacked and the one being tested are the same.} and black-box\footnote{The DeepFake detector being attacked and the one being tested are NOT the same.} attacks, as well as visual fidelity. Our method has a much smaller impact on image quality compared to those adversarial attacks, yet it achieves superior performance in attacking DeepFake detectors. From Table \ref{tab:ff-acc} and Table \ref{tab:celeb-acc} we could observe that the performances of the four adversarial attacks are not stable, \ie,  in many black-box scenarios, the accuracy of images perturbed by other attacks tend to increase. Only CW and PGD could compare with our method in white-box scenarios, but their performance in black-box scenarios is inferior to ours. Jitter and Pixel perform worse than our method in both white-box and black-box scenarios. In contrast, our method consistently interferes with the decision-making of the DeepFake detector, especially showing stable performance in all the black-box scenarios. The stability of our approach may be attributed to the way we add noise. We use Gaussian noising and Gaussian filtering for all the camouflage operations, which are independent of the specific architecture of DeepFake detector. This reduces the risk of overfitting to a particular detector type, making our method inherently detector-agnostic.

Both qualitative and quantitative results strongly prove the superiority of our method in terms of imperceptibility, effectiveness, and transferability.

\begin{table}[!t]
\centering
\caption{SSIM, PSNR and FID scores on FaceForensics++ dataset. }\label{tab:ff-ssim}
\begin{tabular}{ l | c | c | c }
\toprule

Attacks & SSIM$\uparrow$ & PSNR$\uparrow$ & FID$\downarrow$ \\

\midrule
Ours-Xception & 0.99 & 35.38 & 12.22 \\
Ours-FFD & 0.99 & 38.95 & 11.98 \\
Ours-SPSL & 0.99 & 38.93 & 10.87 \\
Ours-SRM & 0.99 & 38.35 & 11.49 \\
\midrule
CW-Xception   & 0.99 & 47.30 & 19.61 \\
CW-FFD        & 0.99 & 46.77 & 18.14 \\
CW-SPSL       & 0.99 & 47.31 & 16.15 \\
CW-SRM        & 1.00 & 50.64 & 2.29 \\
\midrule

Jitter-Xception & 0.84 & 35.54 & 75.90 \\
Jitter-FFD    & 0.84 & 34.46 & 79.19 \\
Jitter-SPSL   & 0.83 & 34.29 & 78.43 \\
Jitter-SRM    & 0.84 & 34.44 & 69.55 \\
\midrule

PGD-Xception  & 0.81 & 33.45 & 91.13 \\
PGD-FFD       & 0.81 & 33.59 & 94.08 \\
PGD-SPSL      & 0.81 & 33.57 & 85.76 \\
PGD-SRM       & 0.82 & 33.74 & 86.52 \\
\midrule

Pixle-Xception & 0.96 & 51.16 & 96.44 \\
Pixle-FFD     & 0.98 & 52.86 & 46.29 \\
Pixle-SPSL    & 0.96 & 51.25 & 89.60 \\
Pixle-SRM     & 0.99 & 53.02 & 40.40 \\
\bottomrule
\end{tabular}
\end{table}

\begin{table}[htbp]
\centering
\caption{SSIM, PSNR and FID scores on Celeb-DF dataset.}\label{tab:celeb-ssim}
\begin{tabular}{ l | c | c | c }
\toprule

Attacks & SSIM$\uparrow$ & PSNR$\uparrow$ & FID$\downarrow$ \\

\midrule
Ours-Xception & 0.99 & 35.73 & 12.92 \\
Ours-FFD & 0.99 & 39.33 & 13.55 \\
Ours-SPSL & 0.99 & 40.01 & 11.39 \\
Ours-SRM & 0.99 & 39.16 & 12.29 \\
\midrule

CW-Xception     & 0.99 & 47.99 & 23.37 \\
CW-FFD          & 0.99 & 48.30 & 19.18 \\
CW-SPSL         & 0.99 & 48.55 & 15.79 \\
CW-SRM          & 1.00 & 50.55 & 3.49 \\
\midrule

Jitter-Xception & 0.82 & 34.52 & 100.25 \\
Jitter-FFD      & 0.81 & 34.47 & 106.38 \\
Jitter-SPSL     & 0.81 & 34.25 & 102.64 \\
Jitter-SRM      & 0.81 & 34.40 & 96.09 \\
\midrule

PGD-Xception    & 0.79 & 33.52 & 117.67 \\
PGD-FFD         & 0.79 & 33.64 & 121.71 \\
PGD-SPSL        & 0.79 & 33.63 & 110.81 \\
PGD-SRM         & 0.79 & 33.74 & 119.54 \\
\midrule

Pixle-Xception  & 0.98 & 51.27 & 70.02 \\
Pixle-FFD       & 0.99 & 51.87 & 40.56 \\
Pixle-SPSL      & 0.98 & 51.45 & 56.98 \\
Pixle-SRM       & 0.99 & 51.94 & 36.70 \\
\bottomrule
\end{tabular}
\end{table}

\begin{figure*}[htbp]
    \centering
    \includegraphics[width=\linewidth]{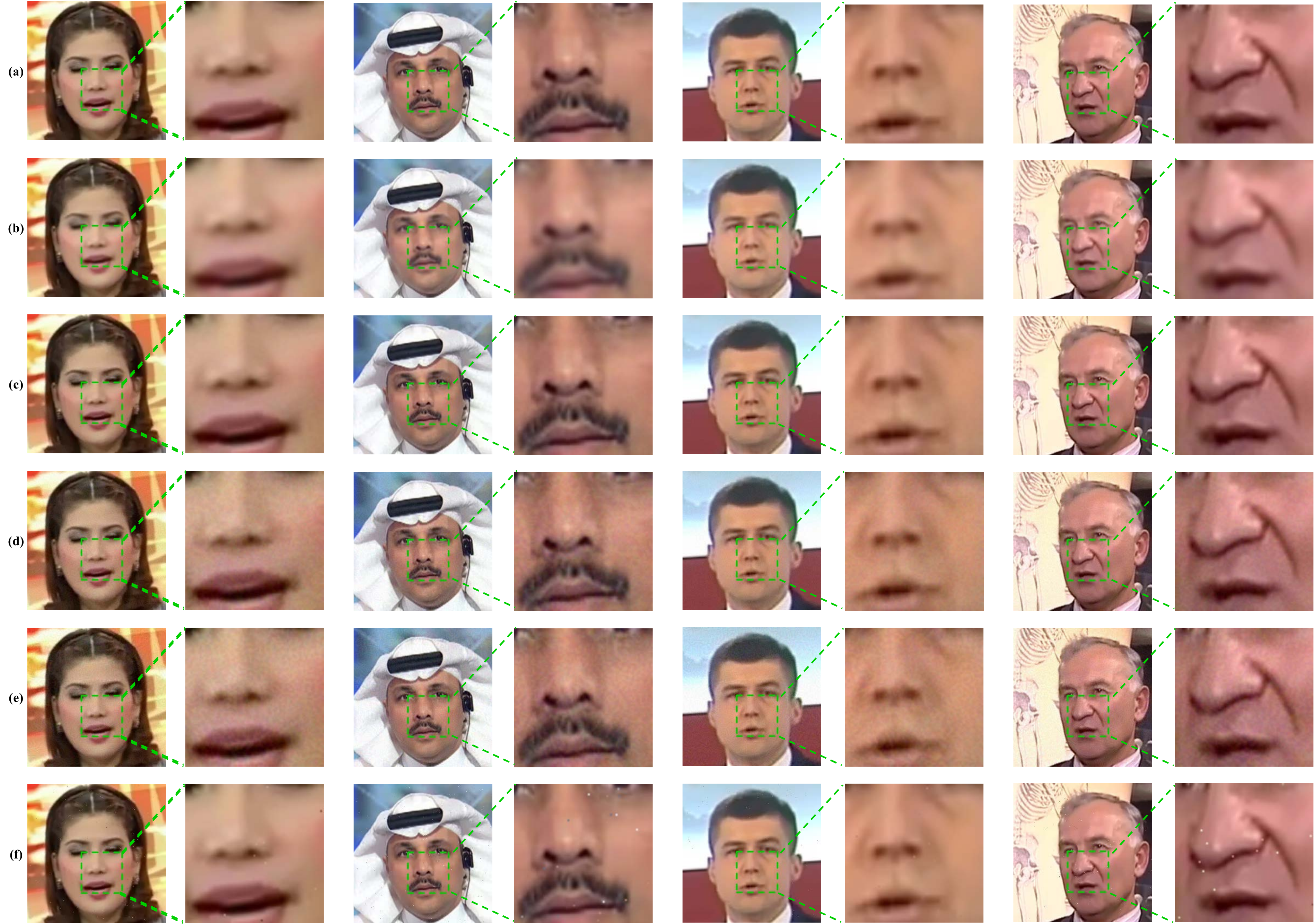}
    \caption{Qualitative Results. We enlarge the areas within the green box for better view. Row (a) represents real clean images. Row (b) represents images camouflaged by our method. Row (c)-(f) represent images attacked by CW, Jitter, PGD, and Pixle, respctively. Note that none of these attack methods could compare with ours in terms of attack success rate. Zoom in to see the details. \red{}}
    \label{fig:04_demo}
\end{figure*}

\smallskip
\noindent\textbf{Qualitative Results.} Row(a) and Row (b) in
Fig. \ref{fig:04_demo} show examples where images are classified as real before camouflage and are classified as fake after camouflage. Row (c)-(f) are images attacked by CW,
Jitter, PGD, and Pixle, respctively. From Row (b) we could observe that images processed by our method have no obvious visual artifacts. Without a close comparison with the real clean face images, it is challenging to discern artifacts in the images from Row (b).  In contrast, the images in Row (d) and Row (e) exhibit grain-like noise(which is typical of adversarial attacks) in the entire image. Although the noise in the images of Row (c) does not appear obvious and the noise in Row (f) consists of sporadic white spots, their attack success rates are not very high either. Additionally, none of these attack methods could surpass our camouflage in terms of attack success rate(See Table \ref{tab:ff-acc} and Table \ref{tab:celeb-acc} for details). We could also observe that in our method, the noise is concentrated only in the facial region, whereas in the images processed by adversarial attacks, noise is distributed across the entire image. The visual quality of our method far exceeds or matches that of the adversarial attacks (Table \ref{tab:ff-ssim} and Table \ref{tab:celeb-ssim}). The main reasons are as follows: 1) Our noise is closely related to facial texture, making it easier to be visually concealed. 2) The noise we add is Gaussian noise and through Gaussian filtering, which, compared to the irregular noise addition of the adversarial attack, appears more natural to the human eyes.

\begin{figure}[htbp]
    \centering
    \includegraphics[width=\linewidth]{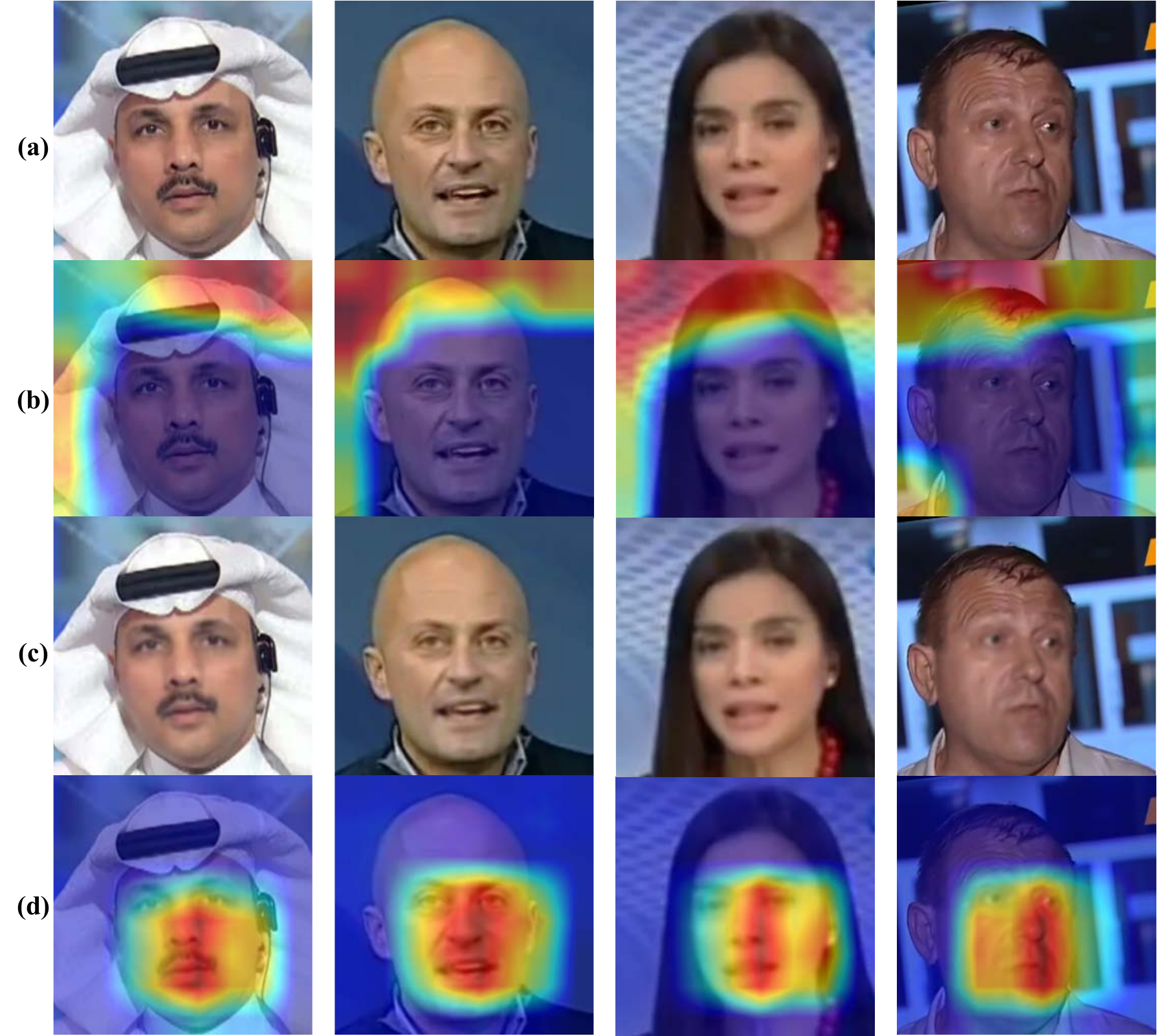}
    \caption{Grad-CAM maps of real images and camouflaged images. Row (a) \& (b): real clean images and their Grad-CAM maps; Row (c) \& (d): Corresponding camouflaged images and their Grad-CAM maps.}
    \label{fig:07gradcam}
\end{figure}

Fig. \ref{fig:07gradcam} shows the different Grad-CAM maps for real clean images and those after being processed by our method. As depicted, our method successfully produce the camouflage described in Sec.~\ref{sec:preval}. The camouflage is subtle enough not to be noticeable to the human eye but effective in deceiving the DeepFake detector. Our method causes the detector to focus on the facial region, producing Grad-CAM maps similar to those in Fig. \ref{fig:06_preval}, ultimately leading to incorrect classification. Our method effectively simulates the effects of DeepFake tampering without compromising the images' quality or any information.

\subsection{Ablation Study}

\begin{figure*}[htbp]
    \centering
    \includegraphics[width=\linewidth]{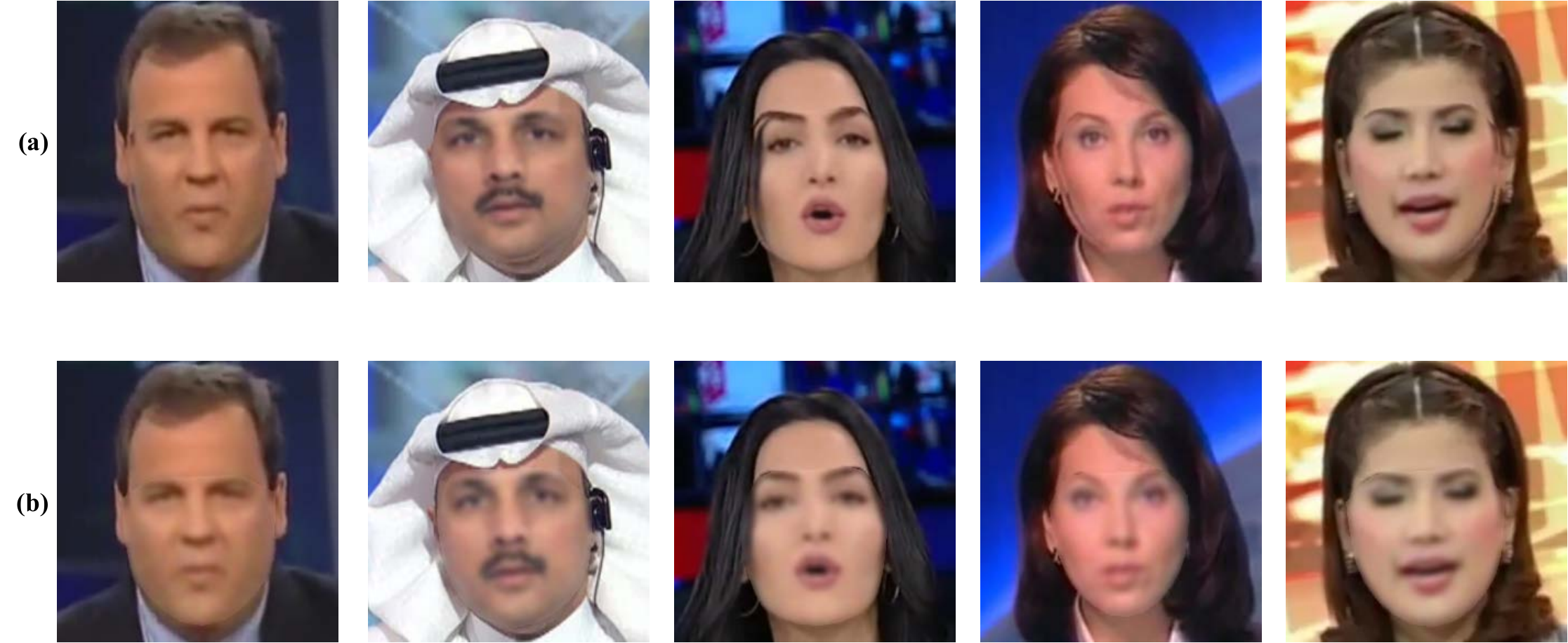}
    \caption{Ablation Study. Row (a) represents images with affine and elastic transforms in camouflage. Row (b) represents camouflaged images without the visual discriminator during training.}
    \label{fig:07_withoutvd}
\end{figure*}

\smallskip
\noindent\textbf{Other operations to create inconsistency.}
Row (a) in Fig. \ref{fig:07_withoutvd} represents camouflaged images with affine and elastic transforms~\cite{chen2022self} in camouflage process. We incorporated affine and elastic transforms into the camouflage process during both training and inference stages, with their essential parameters learned by the configuration generator. As observed in Row (a), Fig. \ref{fig:07_withoutvd}, some images exhibit more pronounced facial edge artifacts than others, reflecting variations in the parameters obtained from different image inputs to the configuration generator. Table \ref{tab:af_el_acc} shows that  after adding the affine and elastic transforms, the performance of our method decreases. Table \ref{tab:af_el-ssim} also indicates a significant degradation in image quality after incorporating affine and elastic transforms. In summary, our camouflage method is already effective in evading DeepFake detectors while preserving image quality well, as demonstrated by both quantitative and qualitative tests.

\begin{table}[htbp]
\centering
\caption{ACC on camouflaged images. FF++ is for FaceForensics++ dataset. AE is for our method with affine and elastic transforms.}\label{tab:af_el_acc}
\begin{tabular}{ l | c | c | c | c }
\toprule

Attacks & Xception  & FFD  & SPSL & SRM\\

\midrule
\midrule
No Attack (FF++) & 0.87 & 0.94 & 0.77 & 0.87\\

\midrule
\midrule
AE-Xception (FF++) & 0.01 & 0.00 & 0.06 & 0.00 \\
AE-FFD (FF++) & 0.06 & 0.00 & 0.10 & 0.04 \\
AE-SPSL (FF++) & 0.09 & 0.03 & 0.14 & 0.06 \\
AE-SRM (FF++) & 0.15 & 0.08 & 0.19 & 0.19 \\
\midrule
\midrule
No Attack (Celeb-DF) & 0.78 & 0.69 & 0.58 & 0.52\\
\midrule
\midrule
AE-Xception (Celeb-DF) & 0.05 & 0.00 & 0.04 & 0.01 \\
AE-FFD (Celeb-DF) & 0.06 & 0.01 & 0.06 & 0.03 \\
AE-SPSL (Celeb-DF) & 0.31 & 0.08 & 0.19 & 0.24 \\
AE-SRM (Celeb-DF) & 0.39 & 0.10 & 0.26 & 0.43 \\
\bottomrule
\end{tabular}
\end{table}

\begin{table}[htbp]
\centering
\caption{SSIM, PSNR and FID scores on camouflaged images. FF++ is for FaceForensics++ dataset.  AE is for our method with affine and elastic transforms.}\label{tab:af_el-ssim}
\begin{tabular}{ l | c | c | c }
\toprule
Attacks & SSIM$\uparrow$ & PSNR$\uparrow$ & FID$\downarrow$ \\

\midrule
AE-Xception (FF++) & 0.93 & 36.11 & 14.48 \\
AE-FFD (FF++) & 0.93 & 36.24 & 14.77 \\
AE-SPSL (FF++) & 0.92 & 36.14 & 15.40 \\
AE-SRM (FF++) & 0.92 & 36.02 & 15.59 \\
\midrule
AE-Xception (Celeb-DF) & 0.94 & 36.23 & 15.14 \\
AE-FFD (Celeb-DF) & 0.94 & 36.22 & 17.04 \\
AE-SPSL (Celeb-DF) & 0.94 & 36.50 & 17.74 \\
AE-SRM (Celeb-DF) & 0.93 & 36.06 & 19.85 \\
\bottomrule
\end{tabular}
\end{table}

\smallskip
\noindent\textbf{Without Visual Discriminator.}
After removing the visual discriminator during the training phase, we utilized the obtained configuration generator to camouflage the images, resulting in images as shown in Row (b), Fig. \ref{fig:07_withoutvd}. It can be observed that after removing the visual discriminator, the noise and blurriness in the facial area of the images become more noticeable, leading to a decline in the visual quality of the images. Furthermore, without the $\mc{L}_\text{vi}$ term to supervise the visual quality, we find that the configuration generator no longer adjusts the attack intensity based on different images but instead applies the maximum degree of camouflage to all input images. This results in almost identical SSIM, PSNR, and FID scores for all methods in Table \ref{tab:nodis-ssim}, which is consistent with our expectations. We test the accuracy of the camouflaged images without the visual discriminator during training and find that they all dropped to $0.0$. This indicates the upper bound of our method's attack effectiveness, \ie, without considering image quality, our method could $100\%$ deceive the DeepFake detectors for all images in our testing sets.



\begin{table}[htbp]
\centering
\caption{SSIM, PSNR and FID scores on camouflaged images. FF++ is for FaceForensics++ dataset.  WD is for our method without the visual discriminator in training. }\label{tab:nodis-ssim}
\begin{tabular}{ l | c | c | c }
\toprule
Attacks & SSIM$\uparrow$ & PSNR$\uparrow$ & FID$\downarrow$ \\

\midrule
WD-Xception (FF++)  & 0.97 & 34.18 & 30.05 \\
WD-FFD (FF++)  & 0.97 & 34.18 & 29.96 \\
WD-SPSL (FF++)  &  0.97 & 34.18 & 30.02\\
WD-SRM (FF++)  & 0.97 & 34.18 & 29.98 \\
\midrule
WD-Xception (Celeb-DF)  & 0.97 & 33.87 & 35.54\\
WD-FFD (Celeb-DF)  & 0.97 & 33.87 & 35.48 \\
WD-SPSL (Celeb-DF)  & 0.97 & 33.87 & 35.53\\
WD-SRM (Celeb-DF)  & 0.97 & 33.87 & 35.46 \\
\bottomrule
\end{tabular}
\end{table}

\subsection{Further Analysis}

\begin{figure*}[!]
    \centering
    \includegraphics[width=\linewidth]{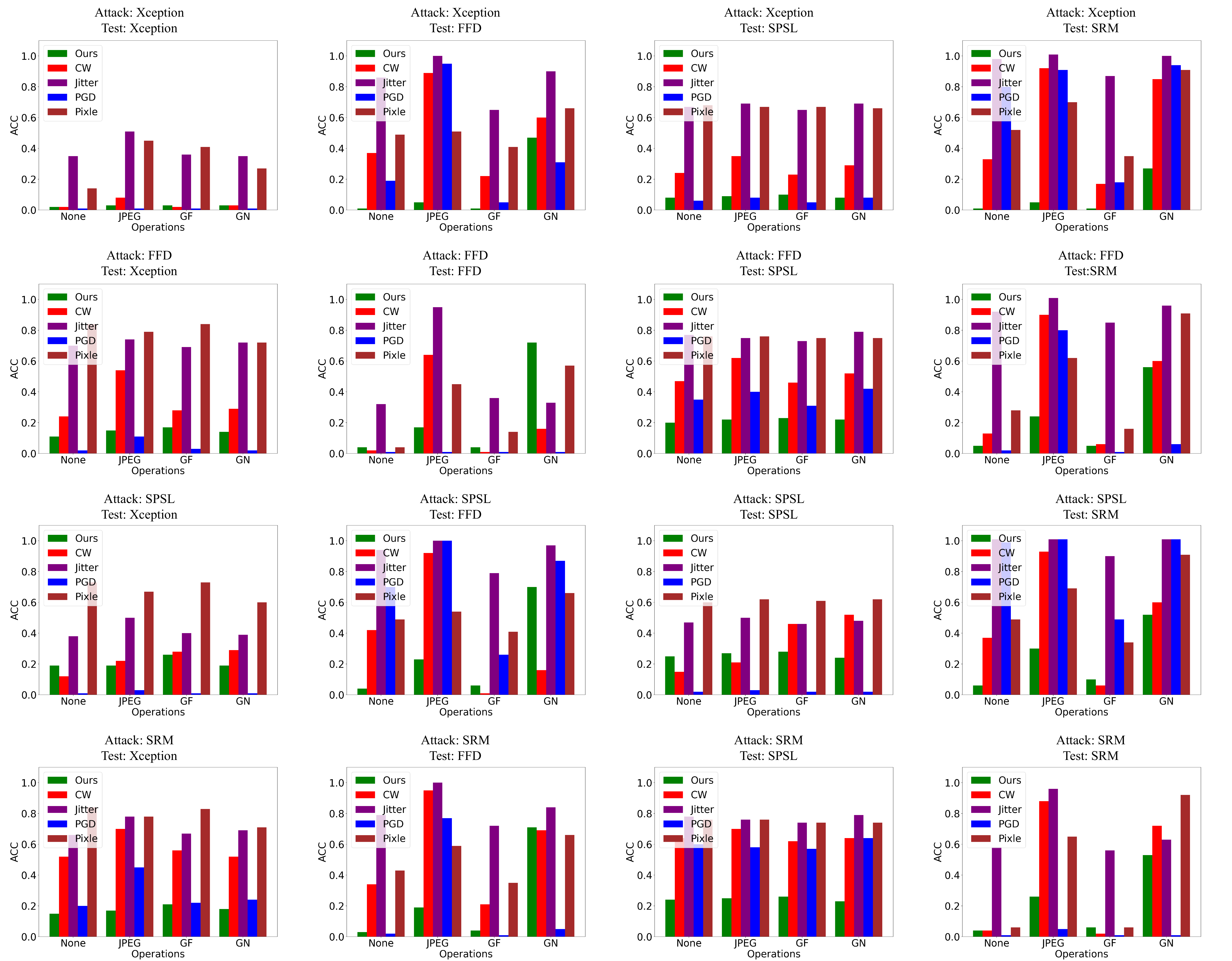}

    \caption{Robustness. None represents attacked images without any post-processing operations. JPEG represents JPEG compression. GF represents Gaussian filtering. GN represents Gaussian noising. We add $0.01$ to all the ACC values for better visualization.}
    \label{fig:08robust}
\end{figure*}

\smallskip
\noindent\textbf{Robustness.}
The robustness of methods is also crucial in real-world scenarios, as images will inevitably undergo various degradations, such as compression during transmission over the 
Internet. To test the robustness our method, we apply post-processing operations to images camouflaged by our method and images perturbed by the four adversarial attacks. Specifically, we use three kinds of post-processing operations, which are JPEG compression (quality factor as $
75$), Gaussian filtering(sigma $0.5$, kernel size $5 \times 5$), and Gaussian noising (sigma $0.01$, mean value $0.0$) respectively. We then test the accuracy of the DeepFake detectors on these post-processed images and compared it to the accuracy on the images without post-processing operations. The results are shown in Fig. \ref{fig:08robust}. The closer the bars are to the X-axis, the lower the accuracy of the images is. Fig. \ref{fig:08robust} shows that our method achieves the best accuracy before and after post-processing and the highest robustness. Although PGD performs slightly better than our method in some cases, overall, our approach is much more stable. In almost all cases, the robustness of our method remains relatively stable without significant fluctuations. Additionally, our method sacrifices far less in terms of visual quality compared to PGD(See Table \ref{tab:ff-ssim} and Table \ref{tab:celeb-ssim}).

\smallskip
\noindent\textbf{Handcrafted Camouflage.}
To demonstrate the effectiveness of our method, we experiment with manually configuring the camouflage parameters. We randomize parameter settings for camouflaging images. The values for $\sigma_\textrm{gn}$, $\mu_\textrm{gn}$, $\sigma_\textrm{gf}$, $k_\textrm{gf}$, $\sigma_\textrm{bl}$, and $k_\textrm{bl}$ are generated completely at random.  As shown in Table \ref{tab:random_acc} and Table \ref{tab:random_ssim}, there is a significant difference in the effectiveness between randomly set parameters and those generated by our trained model, demonstrating  the significance of configuration generator. 
\begin{table}[htbp]
\centering
\caption{Acc on camouflaged images. FF++ is for FaceForensics++ dataset. HC is for handcrafted camouflage.}\label{tab:random_acc}
\begin{tabular}{ l | c | c | c | c }
\toprule
Attacks & Xception  & FFD  & SPSL & SRM \\

\midrule
No Attack (FF++) & 0.87 & 0.94 & 0.77 & 0.87 \\
HC (FF++) & 0.46 & 0.25 & 0.49 & 0.39 \\
\midrule
No Attack (Celeb-DF) & 0.78 & 0.69 & 0.58 & 0.52 \\
HC (Celeb-DF) & 0.47 & 0.14 & 0.43 & 0.31 \\
\bottomrule
\end{tabular}
\end{table}

\begin{table}[htbp]
\centering
\caption{SSIM, PSNR and FID scores on camouflaged images. FF++ is for FaceForensics++ dataset.  HC is for handcrafted camouflage.}\label{tab:random_ssim}
\begin{tabular}{ l | c | c | c }
\toprule
Attacks & SSIM$\uparrow$ & PSNR$\uparrow$ & FID$\downarrow$ \\
\midrule
HC (FF++) & 0.98 & 38.35 & 15.25 \\
HC (Celeb-DF) & 0.98 & 38.67 & 19.66 \\
\bottomrule
\end{tabular}
\end{table}

\section{Conclusion}

This paper describes a new active fake method named DeepFake Camouflage to
evade DeepFake detectors. Specifically, we create and blend imperceptible inconsistency to the facial regions of the real images, making them be misclassified as fake. We design a new generative framework, CamGAN, for creating and blending the inconsistency. We design a strategy based on adversarial learning and reinforcement learning to train the framework. Extensive experiments on the FaceForensics++ and Celeb-DF datasets demonstrate the efficacy and superiority of our method.

\noindent{\bf Acknowledgments.}
This material is based upon work supported by National Natural Science Foundation of China, NSFC No.62271466.

\bibliographystyle{IEEEtran}
\bibliography{ref}

\end{document}